\documentclass[conference, letterpaper]{IEEEtran}
\usepackage{cite}

\ifCLASSINFOpdf
  \usepackage[pdftex]{graphicx}
\else
\fi
\usepackage[cmex10]{amsmath}
\usepackage{algorithmic}

\usepackage{mdwmath}

\usepackage{eqparbox}
\usepackage{url}

\ifCLASSINFOpdf
   \usepackage[pdftex]{graphicx}
\else
\fi

\usepackage[cmex10]{amsmath}
\usepackage{color}
 \usepackage{fancyhdr}

\usepackage{wrapfig}
\usepackage{lipsum}
\usepackage{subfig}
\usepackage{kpfonts}
\newcommand*{\vv}[1]{\vec{\mkern0mu#1}}
\makeatletter
\newcommand{\mypm}{\mathbin{\mathpalette\@mypm\relax}}
\newcommand{\@mypm}[2]{\ooalign{%
  \raisebox{.1\height}{$#1+$}\cr
  \smash{\raisebox{-.6\height}{$#1-$}}\cr}}
\makeatother

\renewcommand{\thispagestyle}[2]{} 
\usepackage{epstopdf}
\AppendGraphicsExtensions{.tif}

\fancypagestyle{plain}{
        \fancyhead{}
        \fancyhead[C]{first page center header}
        \fancyfoot{}
        \fancyfoot[C]{first page center footer}
}
\begin{document}

%
\title{Near Real-Time Data Labeling Using a Depth Sensor for EMG Based Prosthetic Arms}

 \author{\IEEEauthorblockN{Geesara Prathap}
 \IEEEauthorblockA{Faculty of Engineering\\ University of Peradeniya \\
 Email: geesara@pdn.ac.lk}
 \and
 \IEEEauthorblockN{Titus Nanda Kumara}
 \IEEEauthorblockA{The MARCS Institute, \\Western Sydney University\\
Sydney 2750, Australia \\
 Email: T.Jayarathna@westernsydney.edu.au}
 \and
 \IEEEauthorblockN{Roshan Ragel}
 \IEEEauthorblockA{Faculty of Engineering\\ University of Peradeniya \\
 Email: roshanr@pdn.ac.lk}}


%


\maketitle

\begin{abstract}
Recognizing sEMG (Surface Electromyography)
signals belonging to a particular action (e.g., lateral arm raise) automatically is a challenging task as EMG signals themselves have
a lot of variation even for the same action due to several factors.
To overcome this issue, there should be a proper separation which
indicates similar patterns repetitively for a particular action in
raw signals. A repetitive pattern is not always matched because
the same action can be carried out with different time duration.
Thus, a depth sensor (Kinect) was used for pattern identification
where three joint angles were recording continuously which is
clearly separable for a particular action while recording sEMG
signals. To Segment out a repetitive pattern in angle data, MDTW
(Moving Dynamic Time Warping) approach is introduced. This
technique is allowed to retrieve suspected motion of interest from
raw signals. MDTW based on DTW algorithm, but it will be
moving through the whole dataset in a pre-defined manner which
is capable of picking up almost all the suspected segments inside
a given dataset an optimal way. Elevated bicep curl and lateral
arm raise movements are taken as motions of interest to show
how the proposed technique can be employed to achieve auto
identification and labelling. The full implementation is available
at \url{https://github.com/GPrathap/OpenBCIPython}
\end{abstract}


%
\IEEEpeerreviewmaketitle

\section{Introduction}\label{sec:intro}
Developing artificial limbs for amputees is dated back to thousands of years. Up until 15th century, the artificial limbs were made from wood, iron, steel, copper or bronze. They were only passive devices which can be categorized as the most basic type of prosthetics. In the 20th century, with the advancements in electronic technology, scientists were able to produce mechanical movements of the prosthetic arms and legs using motors and drivers. This was the next step and considered as a huge leap of the prosthetic industry. 

Surface EMG (sEMG), which is one of EMG acquiring methods, is more popular among the research community. In sEMG, the input from the muscle activity is acquired from the surface of the skin above the muscle. The current data obtaining technologies and filtering methods are sophisticated enough to remove the interference and crosstalk occurring due to sEMG making sEMG the ideal method for controlling prosthetic devices based on large superficial muscles located near the shoulder, elbow, knee or thigh~\cite{frigo2009multichannel}.

The fundamental problem of the sEMG signal retrieval is the noise. However, when enough samples are provided, machine learning algorithms can learn to ignore the noise presented and train the sEMG model as good as the less noisy iEMG signal based model~\cite{solomonow1994surface}. To accomplish this target, we need to have a large number of labelled training data (if the training process is supervised). However, the labelling process is a tedious process even if the model is trained for several predefined actions or sequences.

The problem of labelling data is significantly prevalent in most of the research conducted. For example \cite{benatti2014analysis}, reported that the sequence the authors have collected had only ten repetitions of a sequence separated by three seconds windows. Further, the authors mentioned that two gestures, open hand and index point had only 75\% accuracy (while the other gestures had 99\% accuracy) because the gestures are not trained enough with different speeds. \cite{bai2016shoulder} reported that they had used 24 sets of data during the experiments. It is obvious that the difficulty of labelling process results in a low number of training data which ultimately leads to a low prediction accuracy. 

\subsection{The intended sEMG based arm controlling mechanism}
Concerning smart or intelligent prosthetics, the researchers favour studying the arm. If the patient is not paralysed, even with a passive prosthetic leg the patient can learn to walk because of the simplicity of the movement and the lower degree of freedom (DOF) of the leg. The real challenge is to control an arm, ideally with movements of fingers. This research conducted by the ``Cambio Wearable Computing Lab", is utilising the Lynxmotion AL5D \cite{lynxmotional5d} 4-DOF robotic arm. The research is based on the following assumptions and hypothesis:

\begin{enumerate}
\item The arm is amputated. Therefore, we cannot record and train the system with sEMG acquired from the surface beneath the shoulder of the arm. All the actions and intended actions are accounted for near and around the shoulder.

\item The arm will be trained to predict the actual angle of the joints, not a predefined action that follows a hard-coded sequence of motions. This moves the problem from classification to regression. The outputs from the regression model are three angles which are shoulder, elbow and wrist that can directly be fed into the robot arm. 

\item The arm will be trained on a healthy subject and will be transferred to an amputee for testing. Since the sEEG signal can significantly differ from subject to subject even for the same action \cite{de1997use} this strategy will result training the system universally a virtually impossible task. Next hypothesis will eliminate this requirement.

\item As long as a generalised sEMG pattern has one to one mapping of each predicted angle of joints and the input and output follow the same locality, we do not have to train the system to suit every user. The user is supposed to learn the movement of the arm through providing required potentials from the shoulder muscles. Therefore, preserving locality will ease the learning process. 

\end{enumerate}

As mentioned above, the training dataset of the system will be required to have a large number of data considering the solution space. All three joints can move nearly 180 degrees of angle. If we assume that a movement will take about 2 seconds and every degree will be recorded, a classic approach is to label the whole two-second window as a particular action. However, in this scenario, the two-second window should be labelled with 180 different angles data so that it can be fed into a regression algorithm. Since the labelling process has to be done for all three joints, in the worst case, it makes 540 labels per two-second window. This is a daunting task and nearly impossible without an automatic labelling method. \emph{This paper presents a method to address this problem, making the labelling process automatic}.


\section{Motivation}\label{sec:motiv}

Most of the EMG-based controlling research suffer from the unavailability of labelled data. Even though, EMG equipment are present, collecting and labelling data has been done manually, which costs a significant amount of time. Since it is time-consuming, the algorithms that were trained has been trained with less amount of input data. If an automatic and real-time data labelling framework is available, the researchers can read as many data as required to feed into the machine learning algorithm. Another advantage of this kind of a system is, the researcher can train the system using all the data available and test on the same system real-time later with entirely new data with added labels. Detecting the accuracy of the system by this method will reduce the overfitting of the algorithm.

Using a third party tool to label the data real-time is an intuitive idea, but surprisingly was never utilised well as someone expected. The closest two approaches were by \cite{blana2016feasibility} where they have utilised a 3D accelerometer, a 3D gyroscope and a 3D magnetometer to model the 3D orientation of an arm on a globe and the second approach was to use a camera and TV-based motion analyser with retro-reflective markers by \cite{frigo2009multichannel}.

We present a simpler method than the current approaches with a single depth sensor (Microsoft Kinect~\cite{kinect}), which specifically is made to track human limb movements for computer games.

\begin{figure*}[!ht]
\begin{center}
\includegraphics[width= 6cm, height= 20cm, angle =90 ]{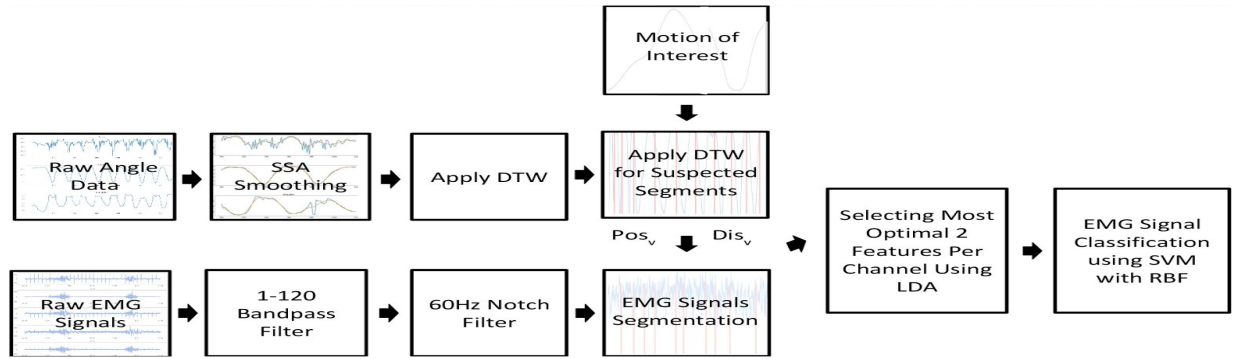}
\caption{\label{f:workflow}The abstract view of the complete system}
\end{center}
\end{figure*}

\section{Related work}\label{sec:Related}

A Kinect-based hand movement detection algorithm was developed by Scherer et al.~\cite{scherer2012kinect}, and both EEG and EMG signals were acquired at the same time for the experiment. Kinect was used to detect two different classes of output, the open hand and closed hand. The capabilities of Kinect was under-utilized because the device was used only to detect two events. Wang et al.~\cite{wang2012human} developed a Human Machine Interface (HMI) using both sEMG and Microsoft Kinect inputs. The architecture is designed to feed the algorithm with either Kinect data or sEMG data from the upper hand to control a human sized service robot. Though the authors have used both sEMG and Kinect, the idea of training the sEMG data with the help of Kinect is not practised. Frigo et al.~\cite{frigo2009multichannel} have conducted an experiment on Multichannel sEMG in clinical gait analysis using a portable sEMG device and multiple cameras to analyse the pattern of sEMG along with kinematic and kinetic of the body. The authors were able to extract the body angles and correlate them with the sEMG data.

To the best of our knowledge, the idea of using a depth sensor such as Microsoft Kinect to label the angles with sEMG data in real-time is not utilised before. This paper will focus on implementation, evaluation and verification of the proposed method. We believe that the proposed solution will help researchers to train a larger number of data samples easily using our framework resulting in a better control accuracy for sEMG based robotic arms.

\noindent{\bf Our Contributions:}\\
Our contributions are, proposing a cheap and easy to implement framework for near real-time angle labeling process. The framework is to be used in sEMG based limb training algorithm which we proposed. We implement the system, evaluate the performance and feasibility of the system by classifying sEMG signals.

\section{Design and Implementation}\label{sec:Design}

The computer connected to the Kinect sensor is sending data as UDP packets while the OpenBCI sensor communicates with the PC using Bluetooth serial communication. Two types of signals are subjected to different pre-processing steps and finally merged into eight channel data streams with five sEMG data and three joint angles. Complete workflow of the system is shown in Fig.~\ref{f:workflow}. Following subsections will explain each component of system in details.

\subsection{Acquiring EMG data from OpenBCI}
The OpenBCI board\footnote{\url{http://www.openbci.com}\label{OpenBCI}} is a consumer grade EEG amplifier based on  Texas Instrument ADS1299 chip that can record various bio-potential signals including EEG, EMG and ECG and proven to be an alternative to medical grade amplifiers\cite{frey:hal-01278245}. OpenBCI has a wireless Bluetooth connection, desktop applications and multiple open source SDKs developed in Python, NodeJS and Java languages. We used an 8-Channel Cyton Biosensing board with Python SDK for the experiment.

\subsubsection{Electrode placements}
\begin{figure}[!h]%
    \centering
    \subfloat[Electrodes placed on 1) Deltoid muscle 2) Pectoralis Major muscle]{{\includegraphics[width=3cm]{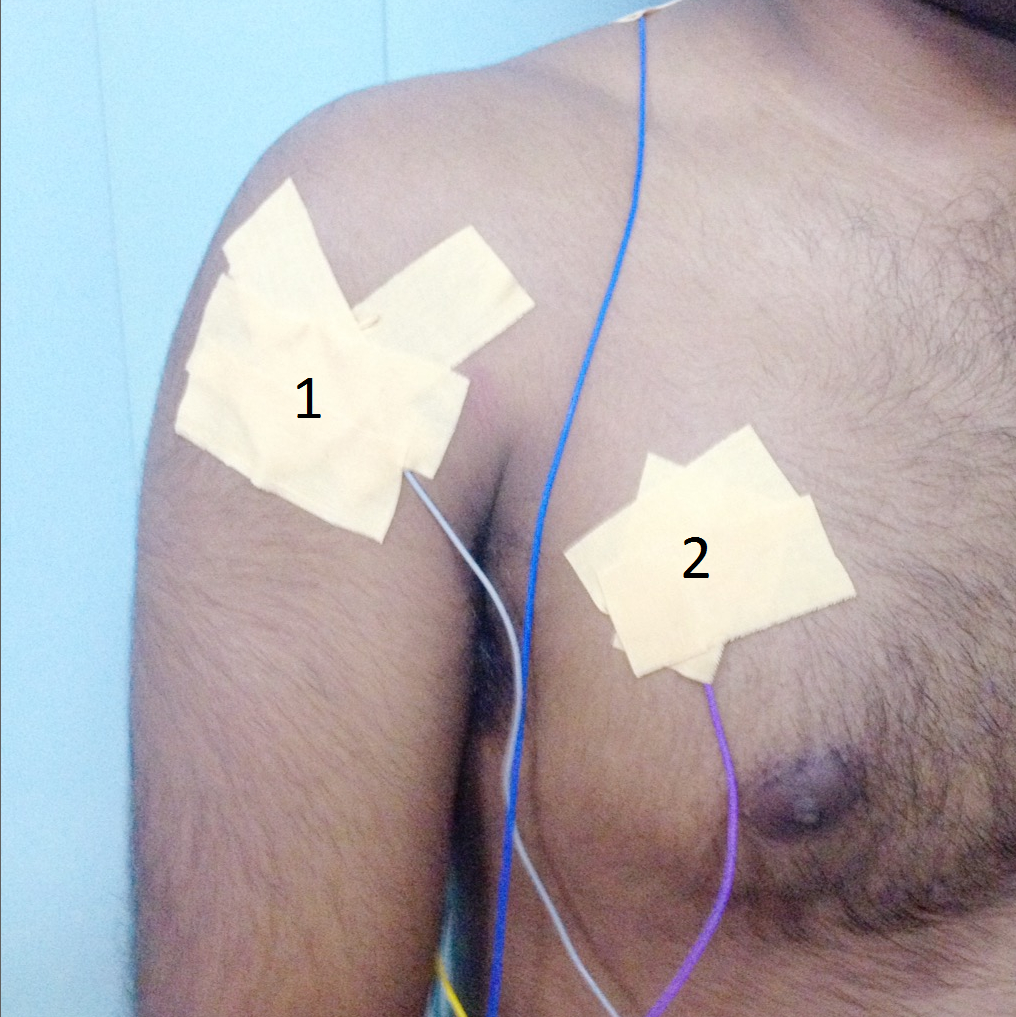}
    \label{f:emgfront} }}%
    \qquad
    \subfloat[Electrodes placed on 3) Trapezius (upper) 4) Trapezius (lower)]{{\includegraphics[width=3cm]{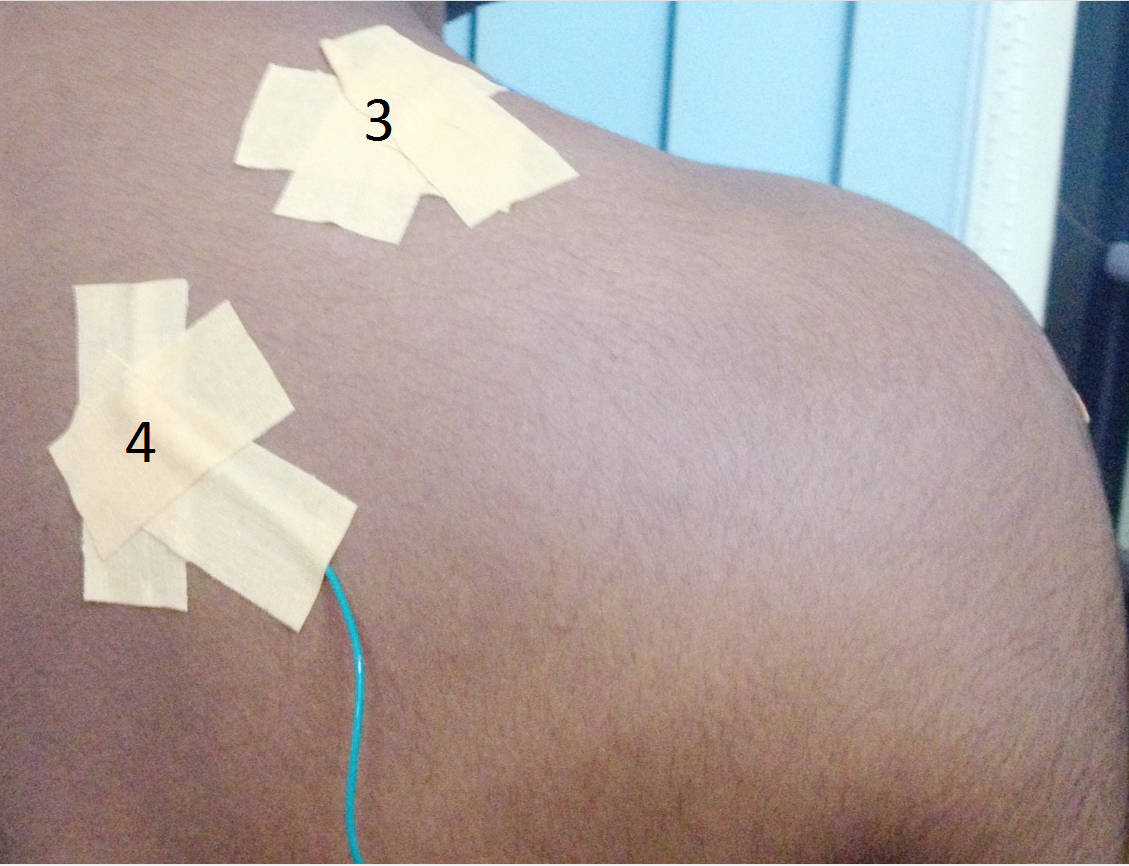} 
    \label{f:emgtop} }}%
    \qquad
    \subfloat[Electrodes placed on 5) Latissimus dorsi muscle]{{\includegraphics[width=3cm]{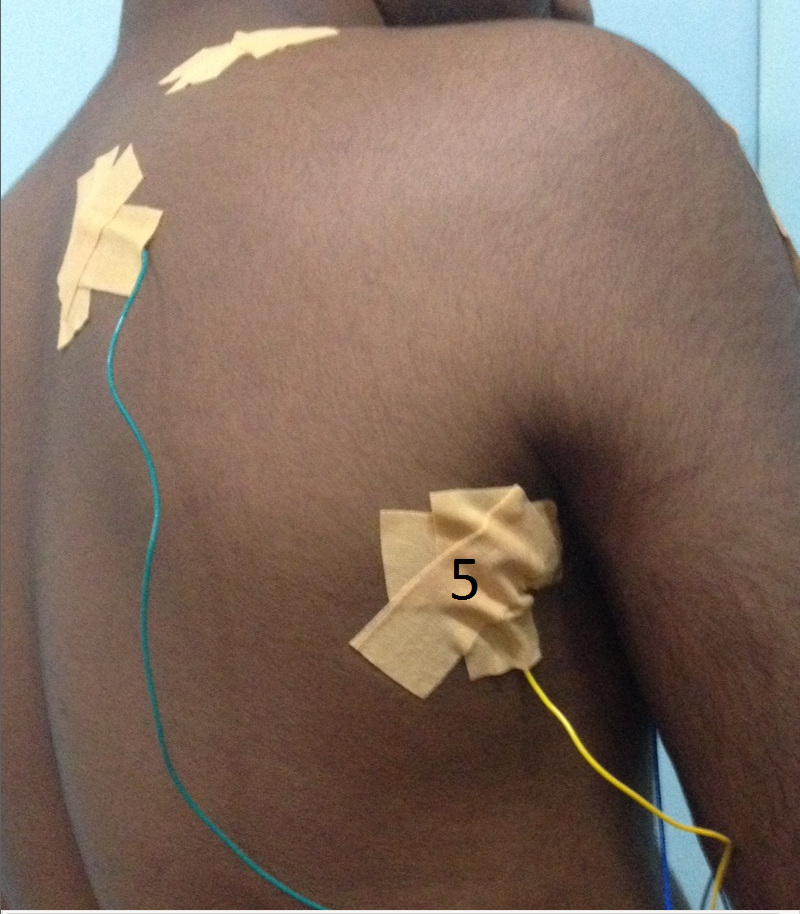} 
    \label{f:emgside} }}%
    \caption{Five electrode placement on major muscles around the shoulder}%
    \label{f:emgelectrodeplacement}%
\end{figure}

Five gold cup electrodes were placed around the shoulder after applying electrode gel and firmly stick on the skin using tape. The selected muscles are: Deltoid muscle, Pectoralis Major, Trapezius (upper), Trapezius (lower), Latissimus dorsi. The electrode placements are shown in Fig.~\ref{f:emgelectrodeplacement}. OpenBCI requires two other electrodes to function properly. They are the ground pin (BIAS pin) and the reference pin. During the experiment, the ground electrode was placed around the wrist of the left hand while the reference electrode is placed near the hip.

\begin{figure}[!ht]
\begin{center}
\includegraphics[width=\linewidth]{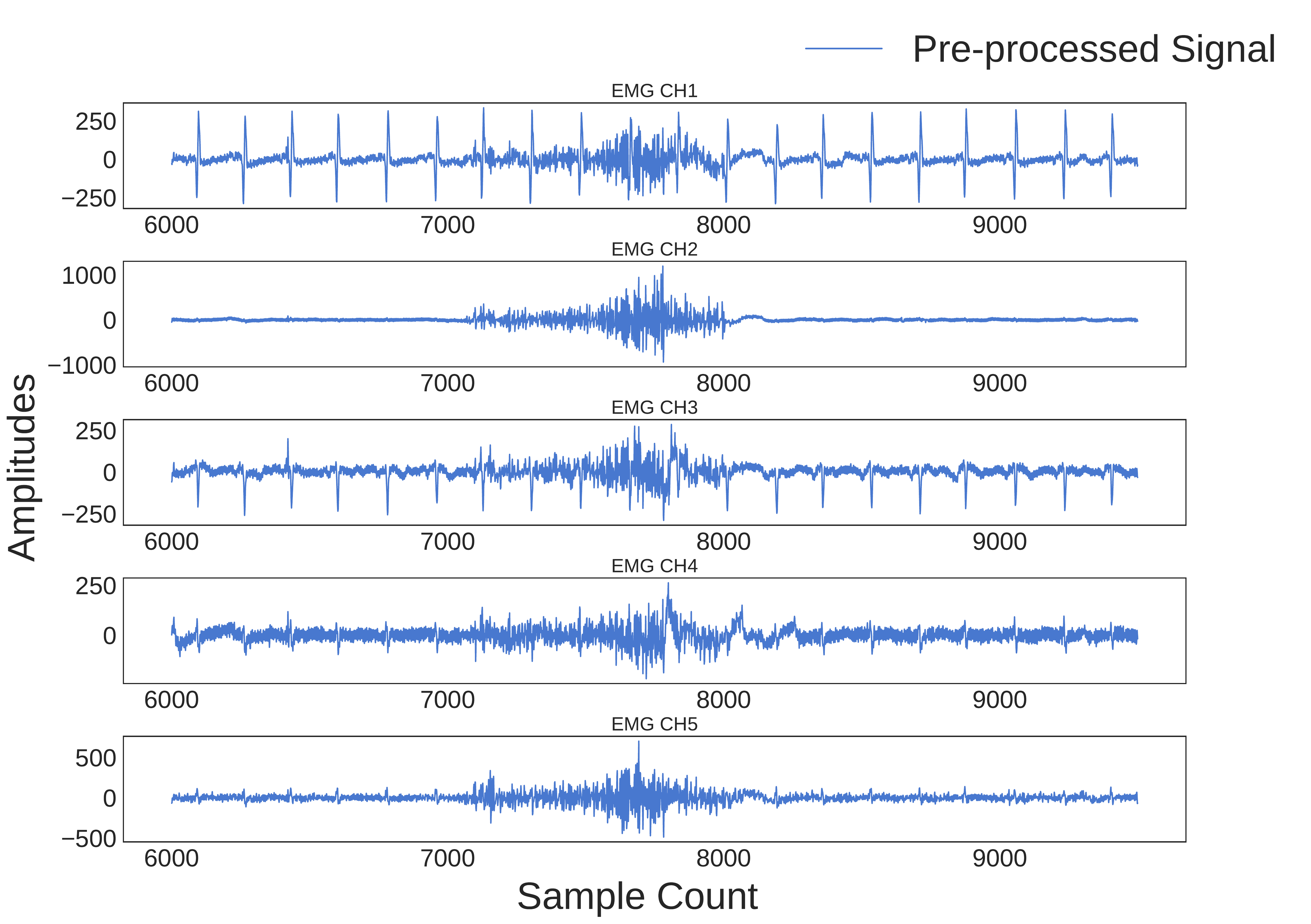}
\caption{\label{f:rawEMGfull}The sEMG measurements from five electrodes after pre-processing.}
\end{center}
\end{figure}

Fig.~\ref{f:rawEMGfull} shows the pre-processed reading from five sEMG electrodes. They are captured from OpenBCI as a raw dump with the sampling rate of 256, then subjected to second order Butterworth low-pass filter of 1-120 Hz to remove high-frequency noise. The recordings are also passed through a 60Hz notch filter to eliminate the interference from 230V AC power line. Fig.~\ref{f:rawEMGfull} contains one iterations of \textit{Elevated Bicep Curl} movement. The periodic pulses frequently appear in the plot are the heart pulses.

\subsection{Microsoft Kinect Sensor}

\begin{figure}[!h]%
    \centering
    \subfloat[Kinect skeleton in seated mode]{{\includegraphics[width=4cm]{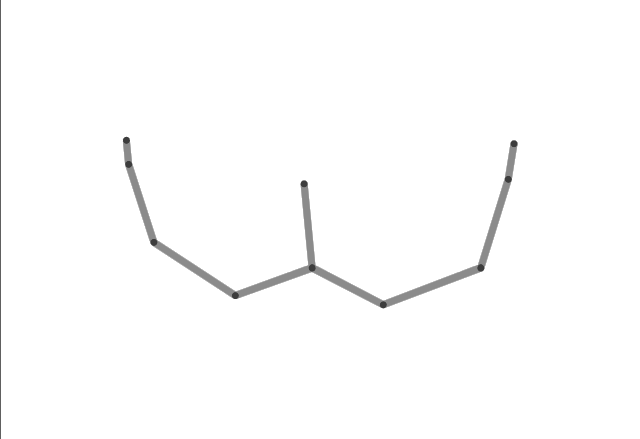} 
    \label{f:seatedkinect} }}%
    \caption{Kinect skeleton in seated mode}%
    \label{fig:kinectOutput}%
\end{figure}

Microsoft Kinect is a sensor which is sold by Microsoft as a part of their Xbox gaming platform~\cite{kinect}. We used the Kinect for Xbox 360 for the experiment. The sensor supports full body 3D motion capturing through an RGB camera and  a Depth Sensor. The skeleton drawing supports two modes, seated mode and default mode. In the seated mode, only the upper half of the body is considered while in the default mode, full body is processed. Since our interest lies only within the upper half, we used the seated mode of the API as shown in Fig.~\ref{f:seatedkinect}.

As visible in Fig.~\ref{fig:kinectOutput}, the Kinect API allows us to access the location of the shoulder, elbow and wrist. Though the API directly do not provide the angles, using the 3D coordinates of joints, it is possible to deduce the angles of each joint.

\subsection{Acquiring joint angles from Kinect}

Kinect describes the bones with respect to the joints. Regarding the right hand, there are four joints that Kinect supports. They also maintain a hierarchy of joints which describe the connections which are equivalent to the bone. The four joints are: right-hand tip(T), right wrist(W), right elbow(E), right shoulder(S).

The Kinect will give the 3D-Cartesian location of each joint (T, W, E, S) by taking the Kinect camera location as the origin. A bone is equivalent to the vector from one joint to the other. For example, the elbow to the shoulder bone can be represented by, $\vv{P} = E-S$
and the elbow to wrist bone can be represented by $\vv{Q} = E-W$
moreover, the angle between two bones is, $angle = \cos ^{ - 1}\bigg( \frac{\vv{P} \bullet \vv{Q}}{\|\vv{P}\|\|\vv{Q}\|} \bigg)$ with four points, three joint angles were calculated for the experiment.

\subsection{Noise Handling}

Data from Kinect contains a considerable amount of noise. The reason is that Body Tracking API of Kinect always tries to model the body angle irrespective of the confidence it has about the position. To make a complete picture in seated mode, both arms should be visible to the Kinect camera. When it is not available, the device is trying its' best to identify both arms resulting wrong positions. However, immediately after both hands are visible, the device returns to the normal state. Apart from this, the device is less sensitive to the small length in between wrist and tip of the hand making the wrist angle measurements noisy.

\begin{figure}[!ht]
\begin{center}
\includegraphics[width=\linewidth]{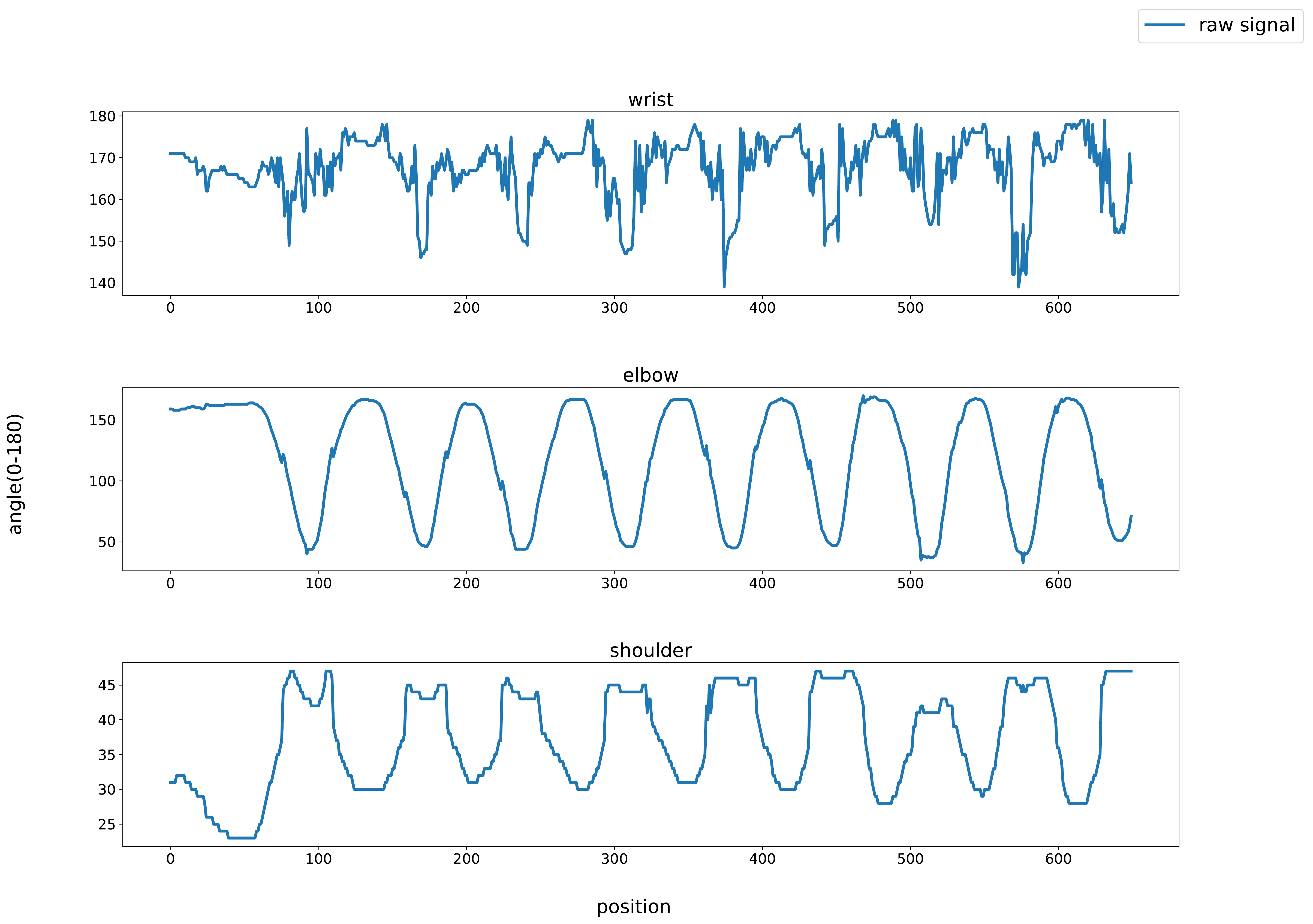}
\caption{\label{f:rawkinectall}Raw angles readings from Kinect for the right arm \textit{Elevated Bicep Curl} movement}
\end{center}
\end{figure}

Fig.~\ref{f:rawkinectall} shows the angle reading for eight \textit{elevated bicep curl} movements. As expected, the Wrist joint shows a significant noise whereas Elbow and Shoulder joint readings are more consistent. To produce a smooth waveform we used Singular Spectrum Analysis (SSA) technique \cite{vautard1992singular}. 

\subsubsection{Singular Spectrum Analysis (SSA)}

SSA is a time series analysis technique, which has many applications. This includes basic tasks such as smoothing, noise reduction, extraction of periodicities and more advanced applications such as Independent Component Analysis (SSA-ICA), time series forecasting and, missing value imputations \cite{golyandina2013singular}. The basic idea behind the SSA can be described as follows.

Let, $X_N = (x_1, \cdots , x_N )$ be a time series of length $N$. A trajectory matrix $X$ can be made from choosing a $L$ amount of data per window, assign the window as a column of the matrix and, shifting the window $K$ times ($K = N-L+1$) where the size of the matrix is $L\times K$. Then the Singular Value Decomposition (SVD) is performed to obtain $U,\Sigma$ and $V$ as $X=U\Sigma V^T$. Then, the matrix $X$ is partitioned as: $X =  \underset{i}{\Sigma}E_i$
where $E_i =  {\sigma}_iU_i{V_i}^T$ is an eigentriple and the index $i$ corresponds to the $i$th singular value ($\sigma$)  or vector ($U$ or $V^T$). Each matrix contains one component corresponding to noise, a trend or oscillatory wave. By choosing a $I$ number of subsets from these eigentriples and then Hankelizing \cite{hassani2007singular} the matrix to original form will produce required feature depending on the choice of the eigentriples. Since our aim is to filter the noise and to smooth the signal, the maximum eigentriple of subset $I$ is chosen and reconstructed as the new time series only by using it \cite{spyrou2014singular}.

\begin{figure}[!ht]
\begin{center}
\includegraphics[width=\linewidth]{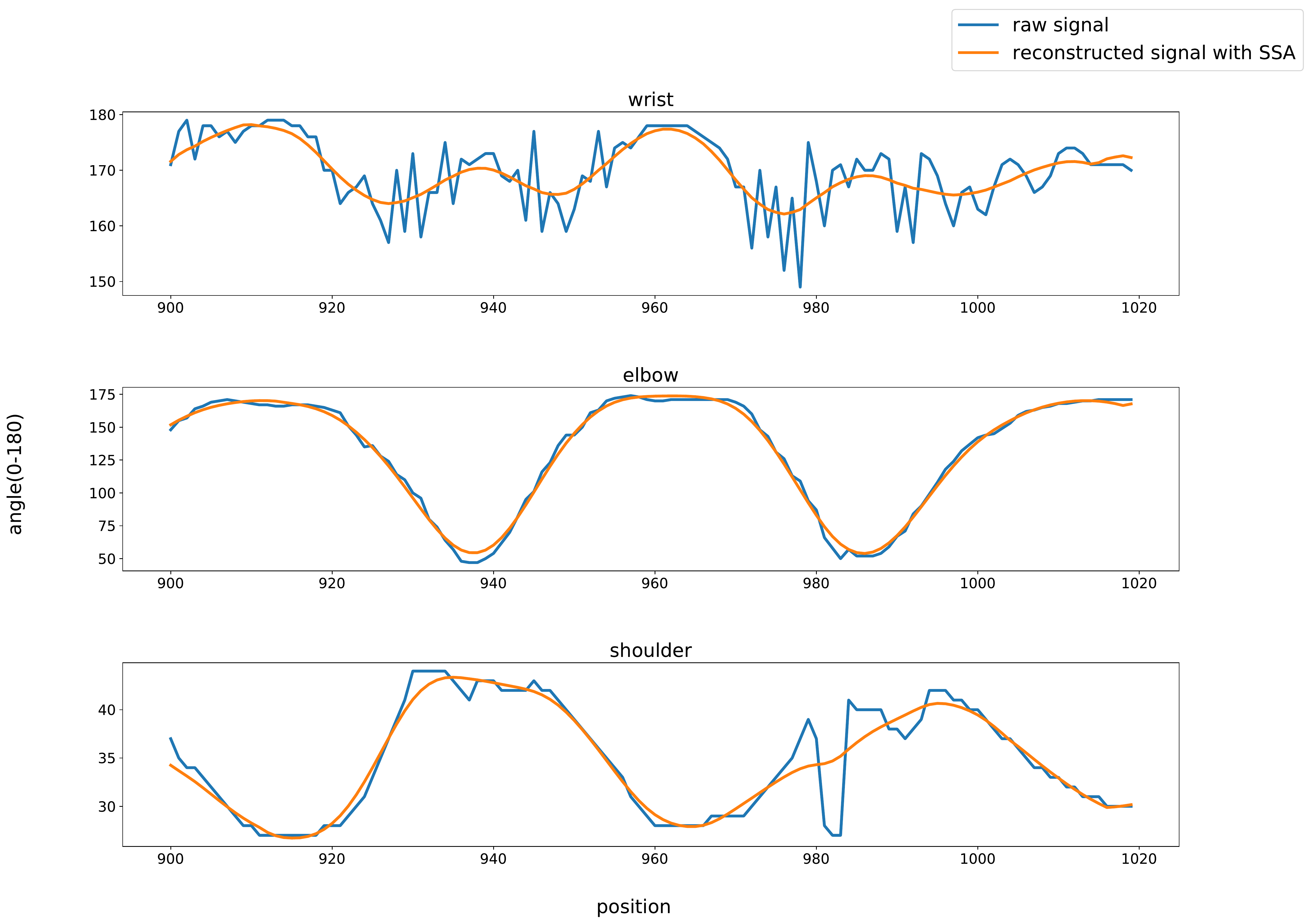}
\caption{\label{f:reconstructodone}A sample of raw signal which is part of Fig.~\ref{f:rawkinectall} for arm movement and the reconstructed signal from SSA}
\end{center}
\end{figure}

Fig.~\ref{f:reconstructodone} shows the reconstructed signal for  \textit{Elevated Bicep Curl} movement. As explained earlier, the SSA technique is a very reliable method of smoothing and noise removal. One of the significant challenges of using a depth sensor like Kinect is noise. The SSA technique can give a proper answer for this problem making the angle measurement more accurate and possible.


\section{Proposed Moving Dynamic Time Warping (MDTW) Technique}\label{sec:method}

For this experiment, we performed two different actions $Act_i \in (a_1,\,a_2)$ where $a_1$ and $a_2$ represent Elevated Bicep Curl and Lateral Arm Raise respectively. The number of repetitive movement corresponding to each action is denoted by $Count(Act_i)$. The following subsection demonstrates how the MDTW technique is employed so as to automate the data labelling process by using Elevated Bicep Curl motion. 

\subsection{Desired Pattern Identification}

\begin{figure}[!h]%
    \centering
    \subfloat[Elevated Bicep Curl Movement]{{\includegraphics[width=3.5cm,height=6cm,keepaspectratio]{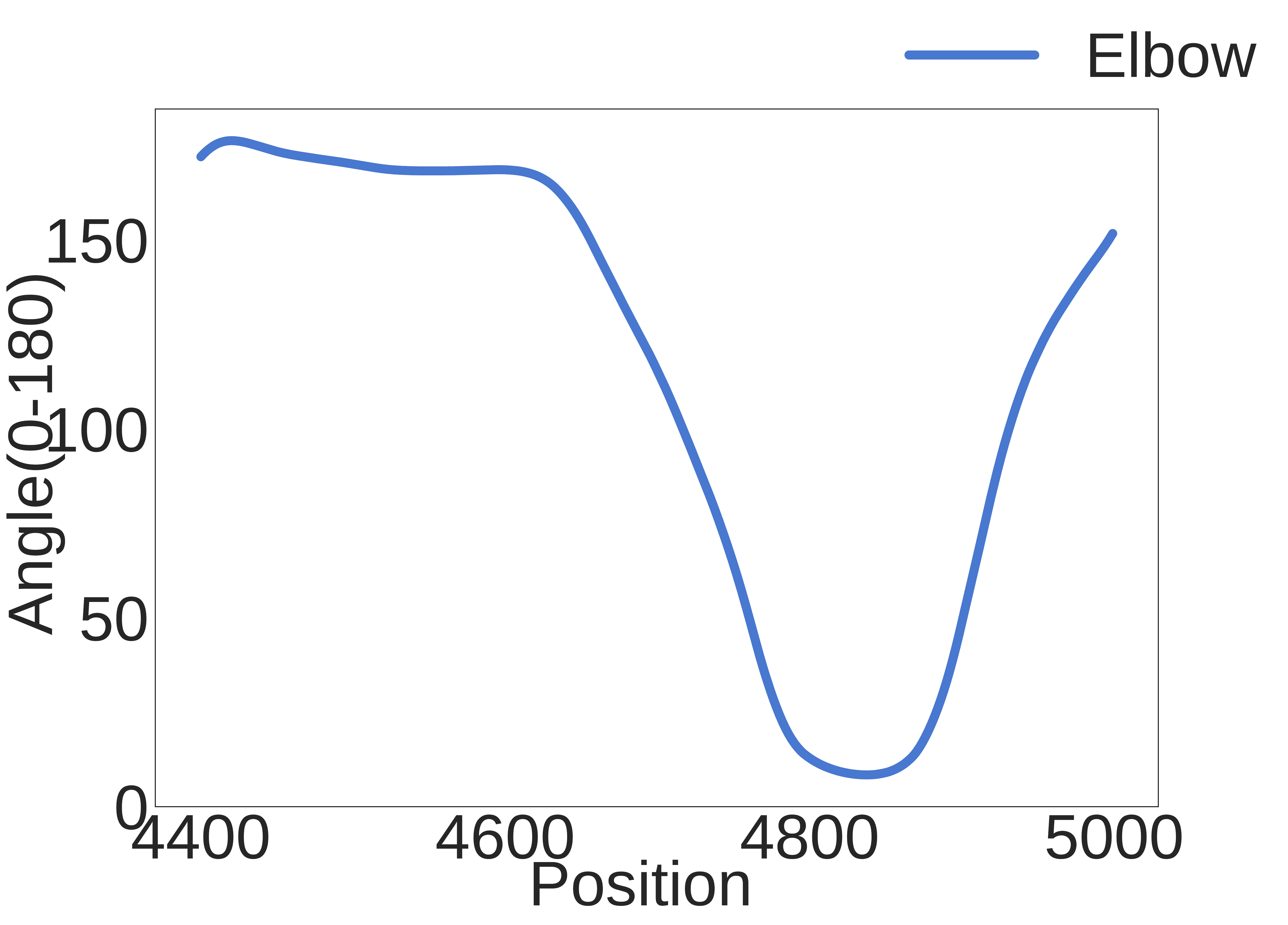}
    \label{f:bicep_motion_of_interest}}}%
    \qquad
    \subfloat[Lateral Arm Raise Movement]{{\includegraphics[width=3.5cm,height=6cm,keepaspectratio]{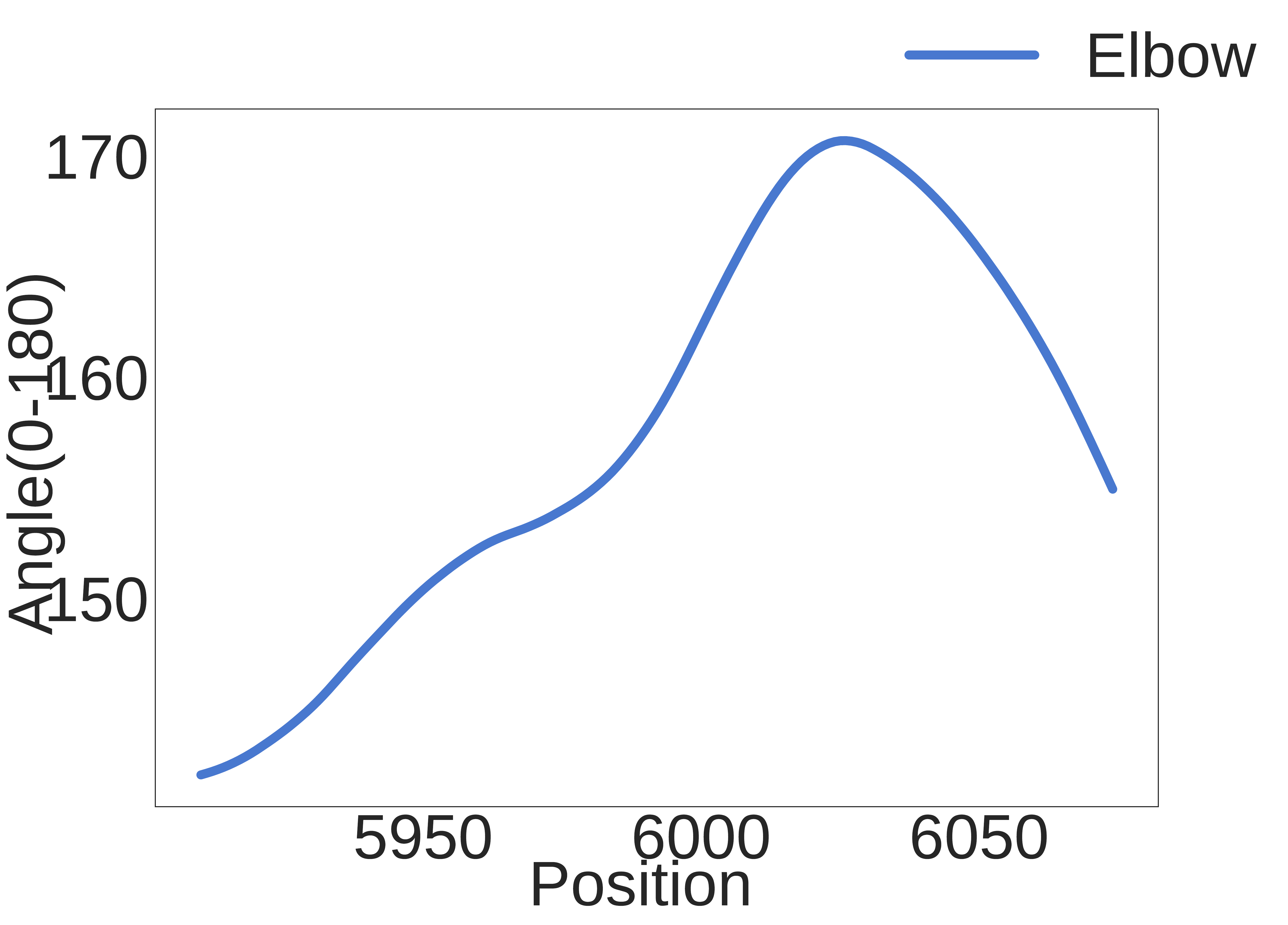} 
    \label{f:straight_up_motion_of_interest} }}%
    \caption{Motion of interest of two actions}%
    \label{fig:resultofchannels}%
\end{figure}

Motion of interest using Kinect for given actions which are shown in Fig~\ref{f:bicep_motion_of_interest} and Fig~\ref{f:straight_up_motion_of_interest} are to be isolated before starting real-time labeling. Isolating sEMG signals itself for a given action is a difficult task. The strength of sEMG signals tends to be highly individualistic due to body type as well as gender. That is why Kinect was employed to identify motion of interest which is independent of the subject as well as the environment. In addition, a motion of interest should be selected such that it represents the average time duration of a particular motion.

\subsection{Apply Moving Dynamic Time Warping (MDTW) on Kinect Angle (elbow)}

\begin{figure}[!ht]
\begin{center}
\includegraphics[width=\linewidth]{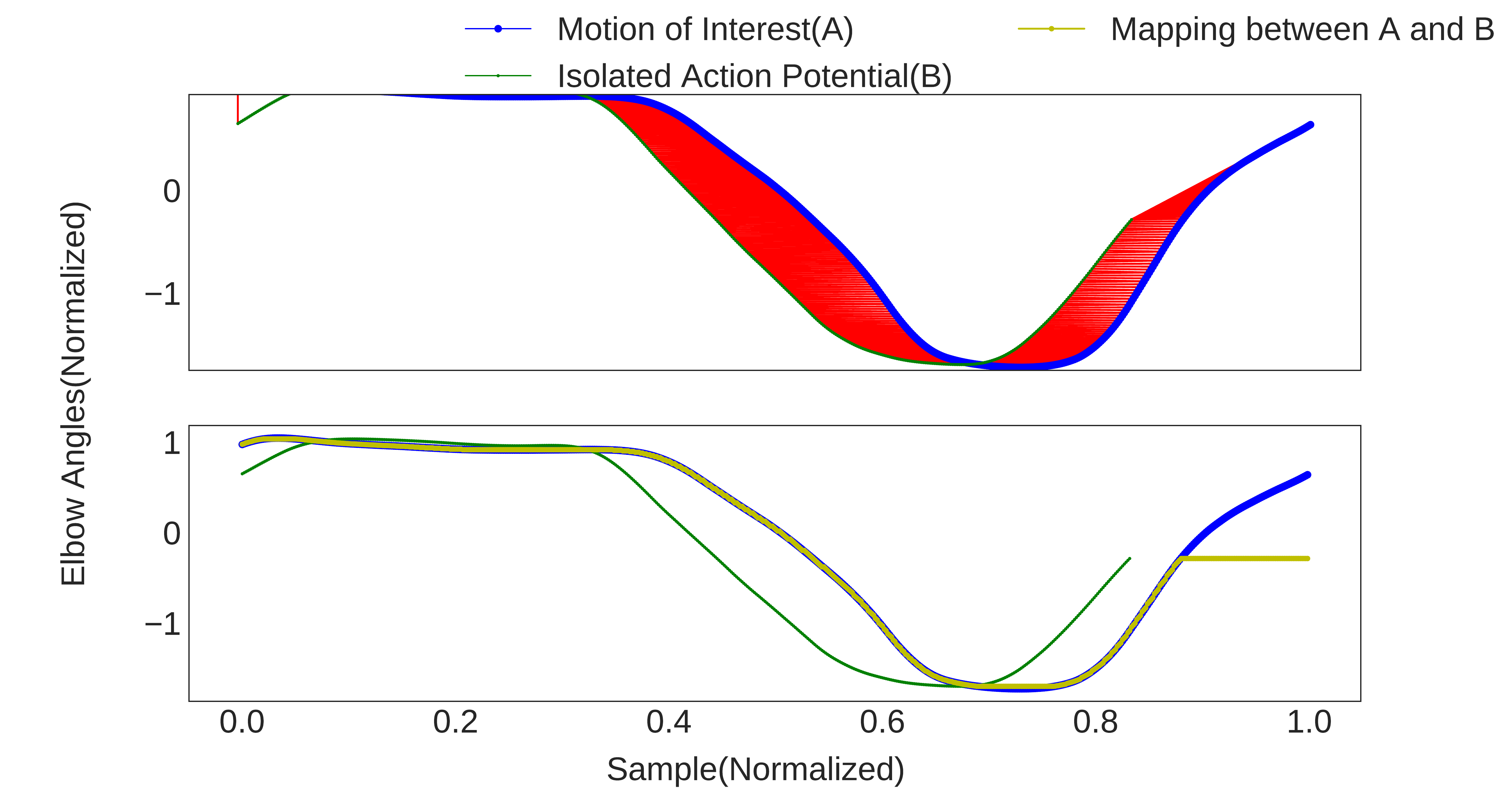}
\caption{\label{f:dtw_mapping} Result of DTW sequence A and B}
\end{center}
\end{figure}

DTW \cite{berndt1994using} is a well-known technique for time series alignment. Let, $A = (a_1,\,\dots,\,a_n) \in \mathcal{R}^n$ and $B = (b_1,\,\dots,\,b_m) \in \mathcal{R}^m$ be two time sequences of interest. This technique can be applied to non-linear mapping between sequences such as A and B in an optimal manner. Let's take $A$ as a motion of interest (Elevated Bicep Curl Movement) and B as a suspected signal. After applying DTW between $A$ and $B$ result can be seen in Fig.~\ref{f:dtw_mapping}. This concept is being used in order to isolate action potentials in this experiment. Width (W) of sliding window is taken twice as long as the length of motion of interest. DTW is applied after collecting first $(b_1,\,\dots,\,b_n)$ data points. After that window moves by one point where DTW is applied to the new window. A distance between the motion of interest and current window is given by DTW. This process is carried out until the end of training where distance vector consists of all the distances and position vector which has indices of corresponding to distance vector is being populated gradually. 

\subsection{Detecting Local Minima of Distance Vector}

\begin{figure}[!ht]
\begin{center}
\includegraphics[width=\linewidth]{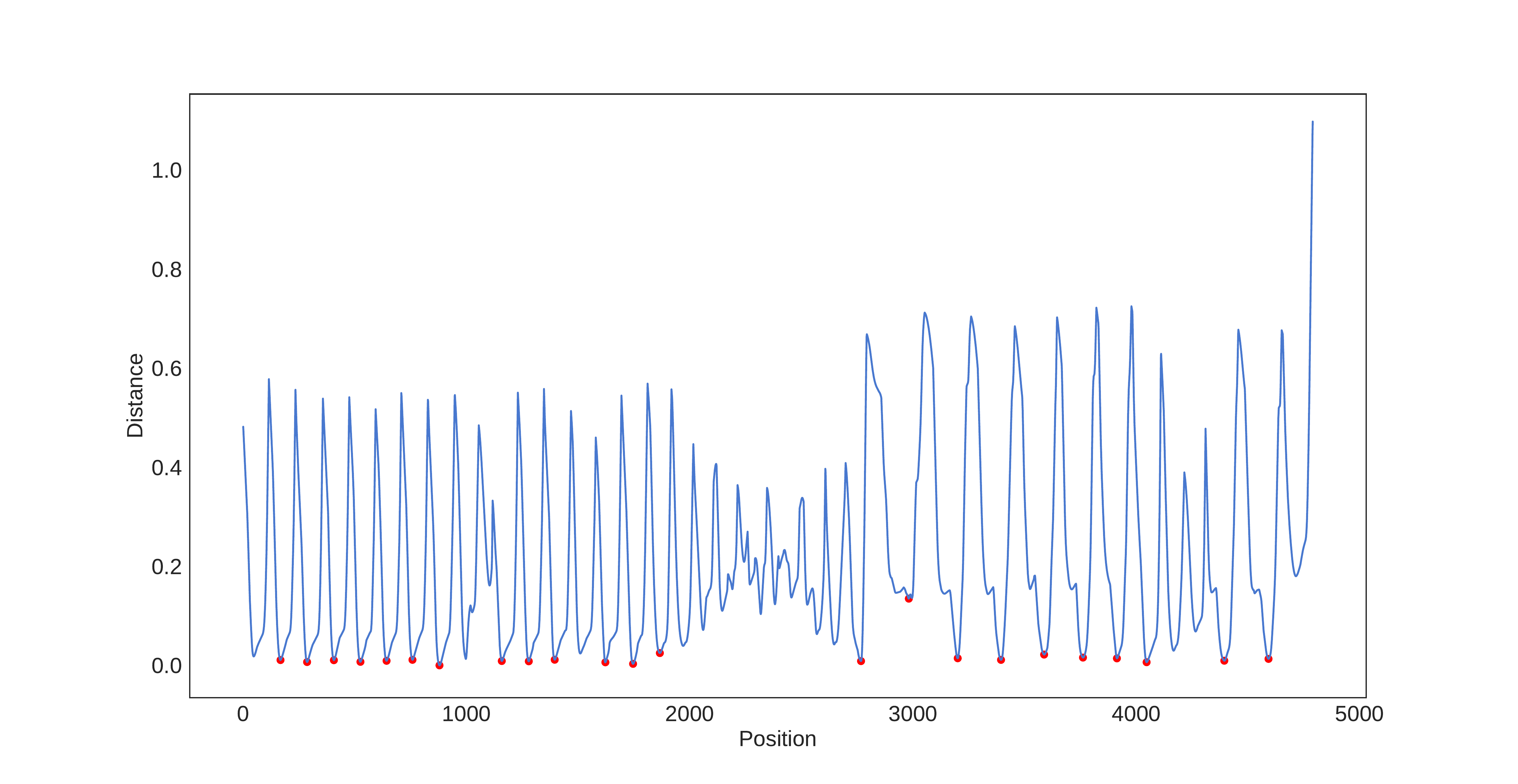}
\caption{\label{f:local_minima_with_0_5}Local Minima Detection with peek threshold 0.5}
\end{center}
\end{figure}

\begin{figure}[!ht]
\begin{center}
\includegraphics[width=\linewidth]{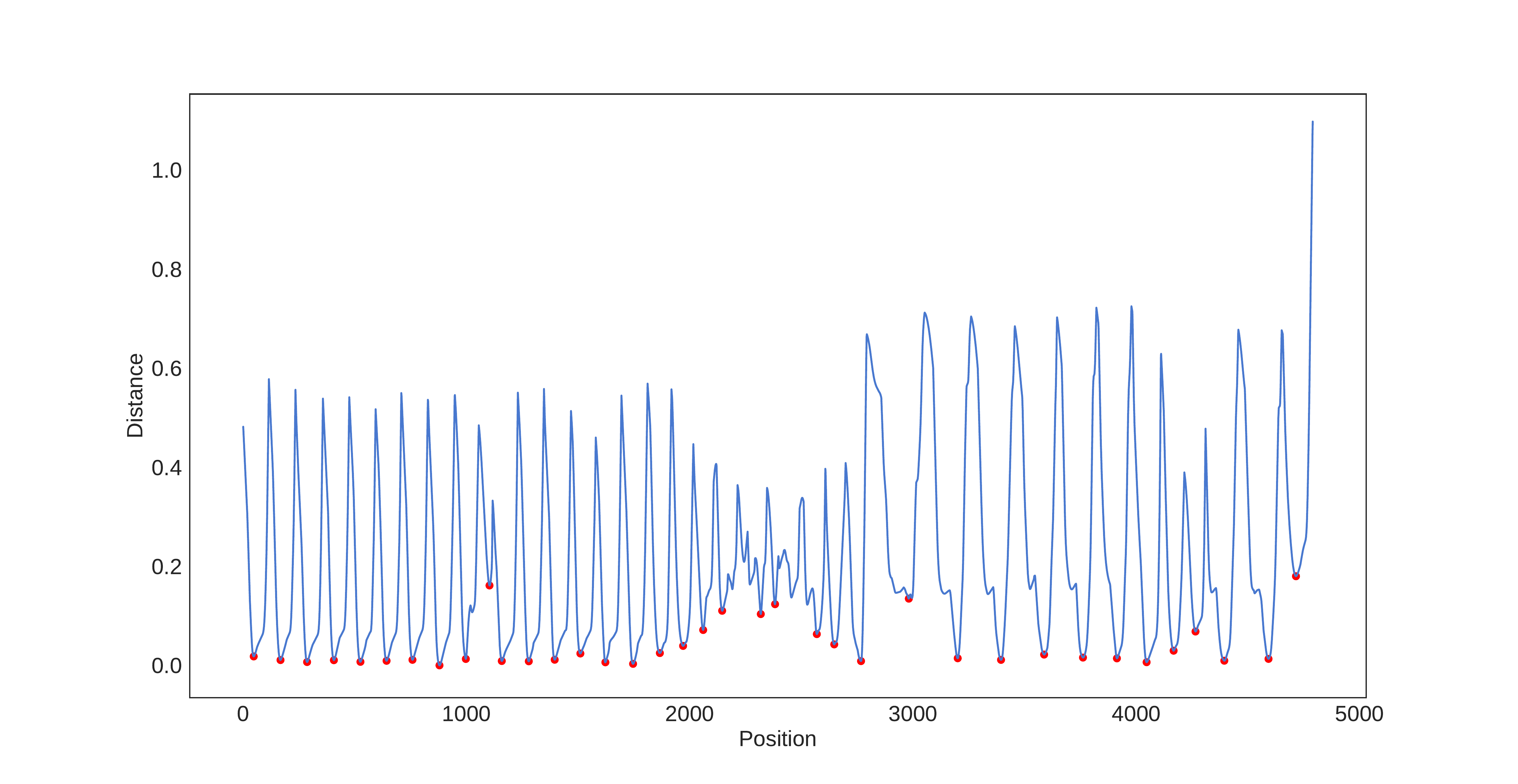}
\caption{\label{f:local_minima_with_recursive}Local Minima Detection with recursive approach}
\end{center}
\end{figure}

If the signal has a lot of noise, detecting local minima of distance vector is a difficult task. Since signal was smoothened using SSA, unexpected false positive can be mitigated. When finding local minima, at least 0.5 (peak threshold) minimum amplitude difference between a peak and it’s surrounding should be required, in order to declare it as local minima. Peak threshold cannot be hardcoded as 0.5 or some value which is closer to 0.5. As it won’t detect all of the local minima. So in here, start with peek threshold as 0.5 and find undetected local minima recursively by applying the same algorithm for each separated sections from the previous result reducing peak threshold dividing by its half. This recursive process is continued up to three level at max. According to Fig.~\ref{f:local_minima_with_0_5}, there are some local minima which were not detected properly. However, after applying in a recursive manner up to 3 level deeper it is capable of detecting almost all the local minima. It can be clearly seen by observing Fig.~\ref{f:local_minima_with_recursive}


\subsection{Extracting Labeled Samples}
\begin{figure}[!ht]
\begin{center}
\includegraphics[width=\linewidth]{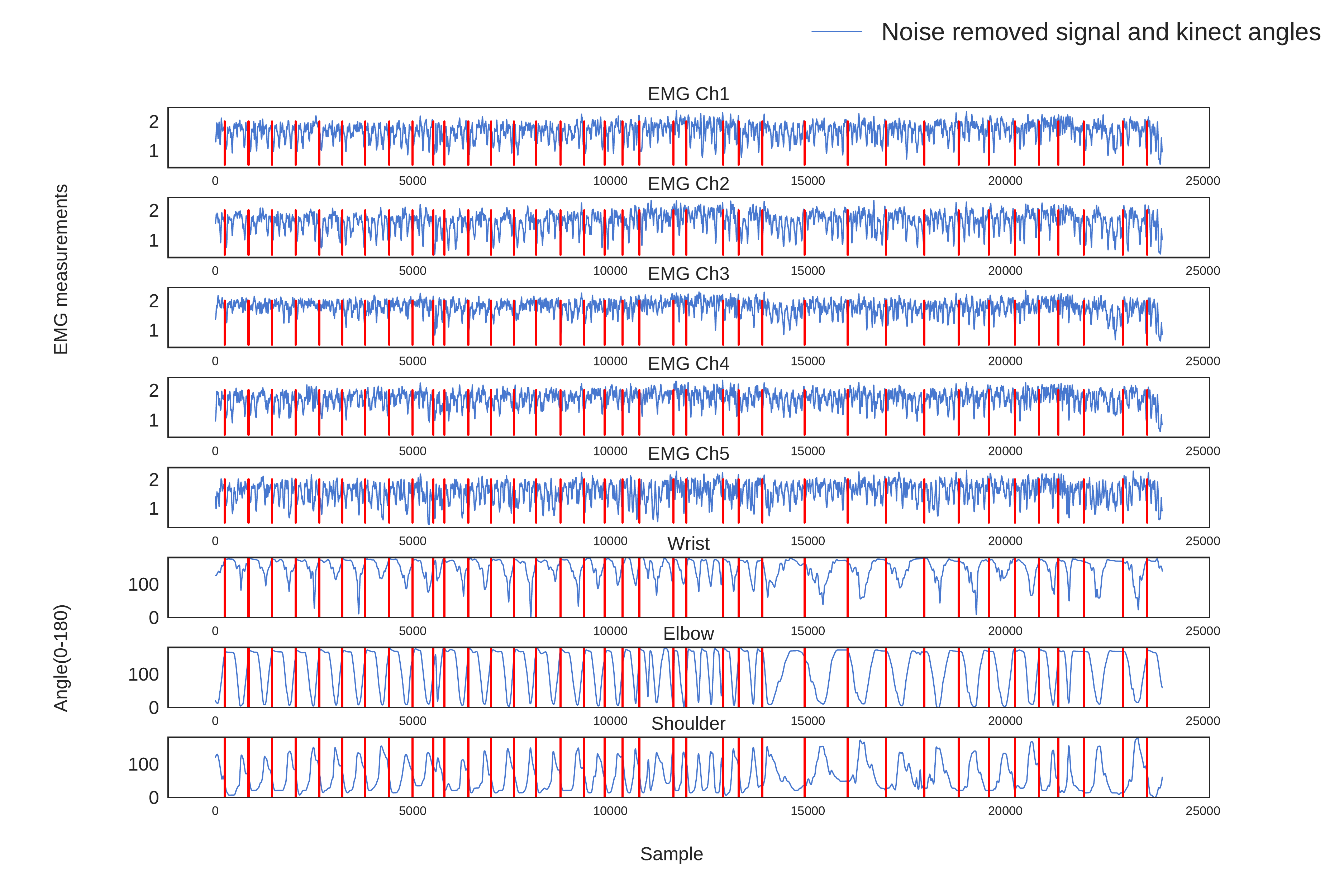}
\caption{\label{f:dwt_result} Labeled data set using Automatic Labeling Algorithm for Elevated Bicep Curl movement}
\end{center}
\end{figure}

After detecting local minima, positions corresponding to local minima can be retrieved from the position vector ($Pos_v$). Along with that suspected signal segmentation can be performed by retrieving temporal signal from joint Kinect angle (elbow) corresponding to an adjacent position of $Pos_v$. DTW is employed on each segment as a one-to-one mapping between a motion of interest and suspected signal. New distance vector ($Dis_v$) is populated by collecting distance of each segment and it needs to be sorted in ascending order. Since number of repetitive actions of a particular action ($Act_i$) is known, $Count(Act_i)$ of segments can be retrieved by using $Pos_v$ and $Dis_v$. Since exact count ($Count(Act_i)$) which corresponds to a particular action is known, corresponding sEMG signals can be segmented and labelled at this stage. Same steps (5.1 to 5.4) need to be applied for each action in a parallel manner. Fig.~\ref{f:dwt_result} shows the final labelled data set for Elevated Bicep Curl movement. 

\section{Evaluation Procedure, Experimental Results and Analysis}\label{sec:Experiment}

To test the proposed technique, obtained sEMG data along with joint angles at the same time corresponding to each action ($Act_i$) which at least about 1-2 seconds. We were able to collect 50 recordings. The duration of all the experiment at least for less than 5 minutes and a total of 100 actions was recorded. The main aim of this project is assessing how well the proposed technique of data labeling performs for 100 recordings of actions, each having 50 have decided enough, because if 50 records can be separated properly, the same technique can be applied when more data is available. Then labeled data is classified into two difference classes to evaluate the proposed technique.

\subsection{Feature Selection and Feature Extraction}

Feature selection and the most suitable feature set extraction are the next challenging tasks. Final classification decision solely depends on this. Even though the main concentration of this project is not regarding feature selection, some attention was given to select and extracting proper feature set. Selected feature set should be represented the best characteristics of unprocessed signals. In this experiment, signals were acquired from 5 different channels. Time domain and frequency domain features were selected by considering the following facts. Random noises: frequency characteristics might change over time. Signals are not stationary: the characteristics of statical properties can be considered for a particular temporal sample only. Following features were selected: median frequency (MDF)~\cite{chowdhury2013surface}, root mean square (RMS)~\cite{daud2013features}, zero crossing rate (ZCR)~\cite{al2016distance}, Willison amplitude (WA)~\cite{chowdhury2013surface}, power spectrum density (PSD)~\cite{hermens1992median}, slope sign change (SSC)~\cite{al2016distance}, spectral centroid (SC)~\cite{mahaphonchaikul2010emg}, probability density function (PDF)~\cite{al2016distance}, spectral entropy (SE)~\cite{vakkuri2004time}, svd entropy (SVD)~\cite{zhao2006emg}. Feature normalization is the next step. Since there is no clue about the whole distribution, standard normalization techniques such as min-max scaling, standardization which first subtract mean and divided by it's variance cannot be applied for temporal sample separately one by one. So in here, log filter was applied by separating out each labeled dataset. \[normalizedLabeledDataset = \log\bigg( {dat[ch_i][fid_j]} \bigg) \] for distinct $ch_i\in \{1,\dotsc,5\}$,   $fid_j\in \{1,\dotsc,10\}$, where labeled sample is denoted by bdat, $ch_i$ is the channel number and $fid_j$ is the feature id.   

There should be an optimal way of selecting some of the features mentioned above which have more bias on final result for a given channel. A selected feature set for channel 3 might not be best suited for channel 5. Likewise, there should be some feature reduction techniques needed to be applied so as to filter out a proper feature set for corresponding sEMG channel. There are different approaches that can be used for feature reduction. Linear discrimination analysis (LDA)~\cite{izenman2013linear} was employed in this experiment to chose most optimal two features per channel.

\begin{figure}[!h]%
    \centering
    \subfloat[Channel 03]{{\includegraphics[width=3.5cm,height=6cm,keepaspectratio]{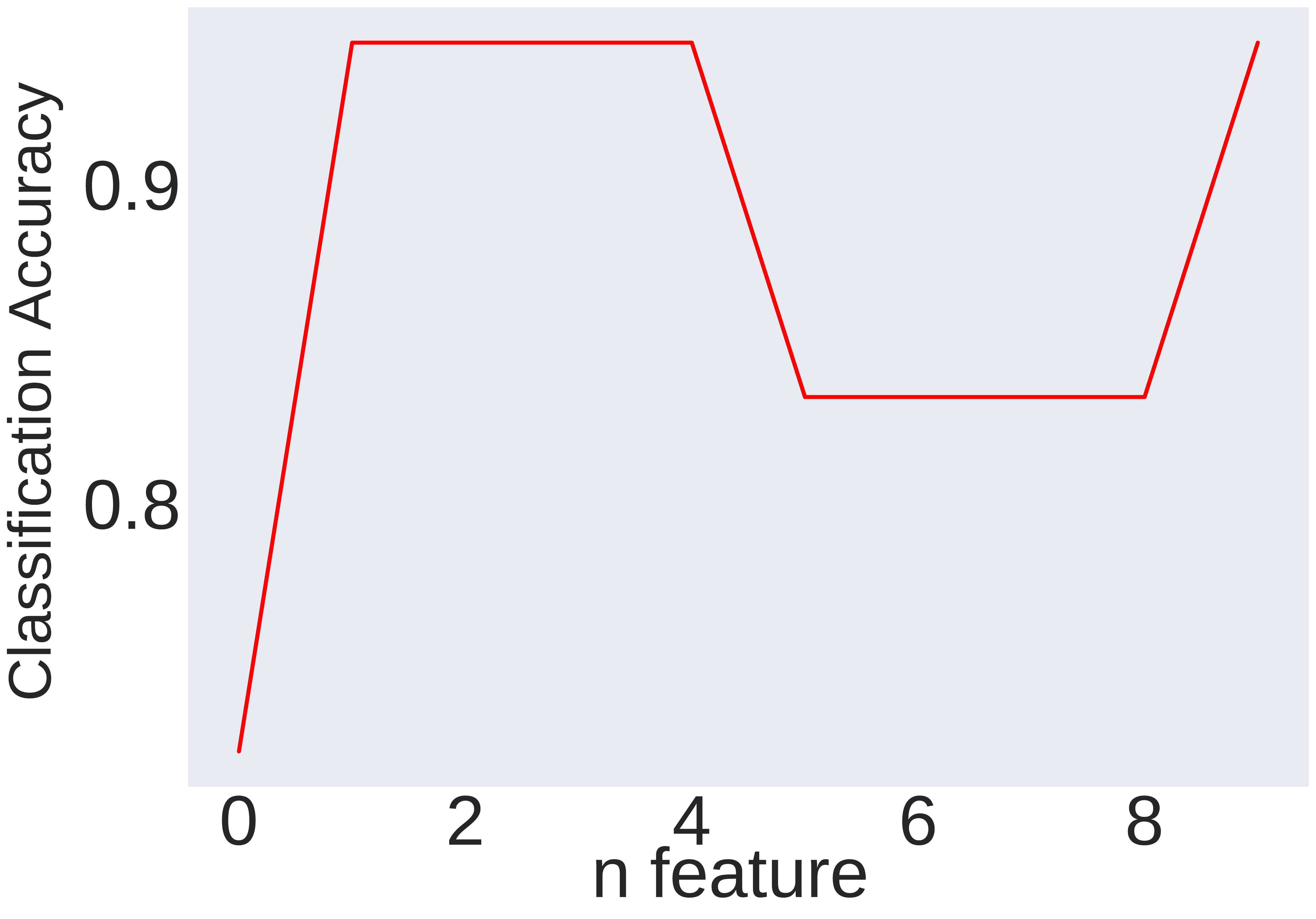}
    \label{f:channel_3} }}%
    \qquad
    \subfloat[Channel 05]{{\includegraphics[width=3.5cm,height=6cm,keepaspectratio]{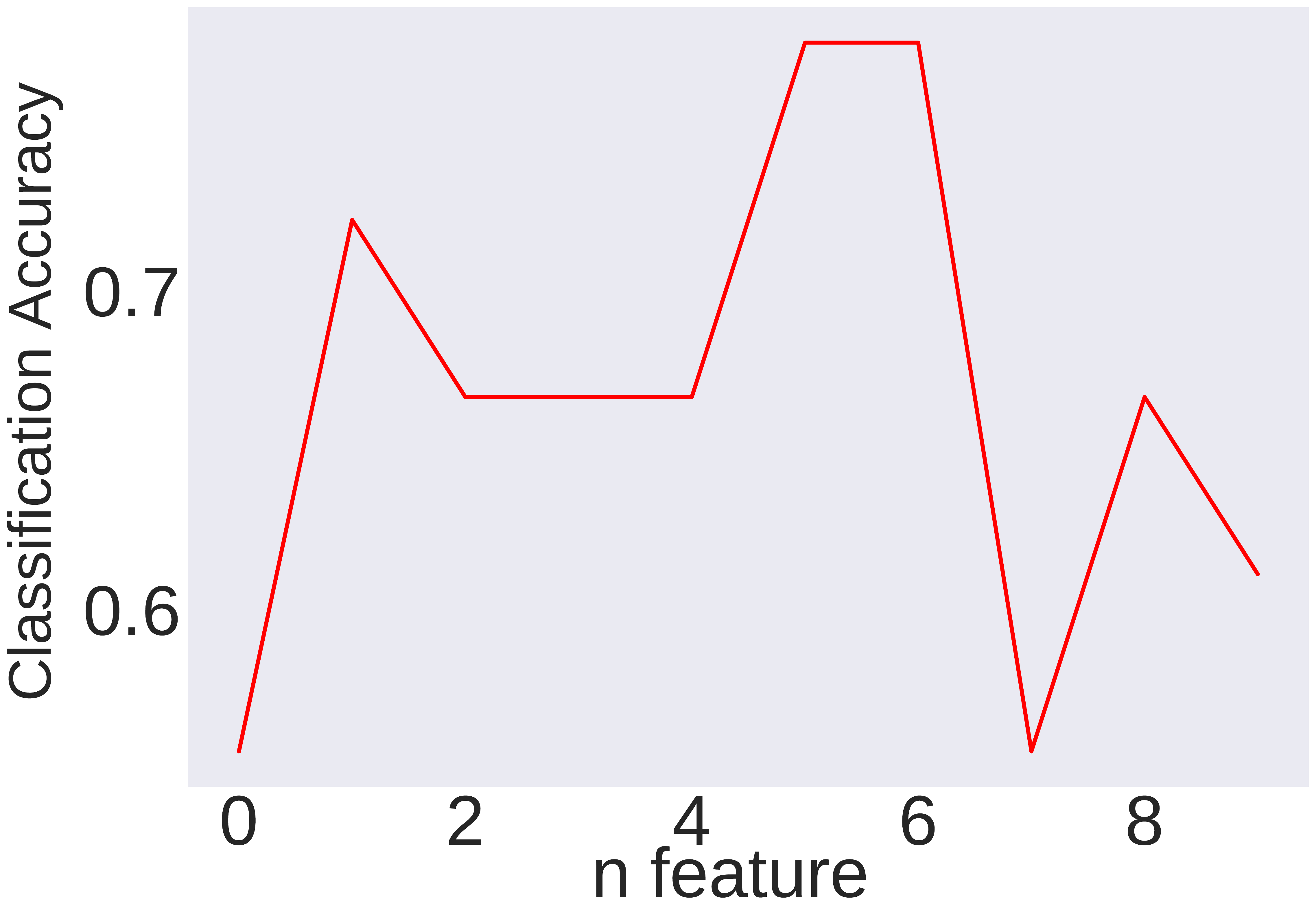} 
    \label{f:channel_5} }}%
    \caption{Classification Accuracy for corresponding features using LDA}%
    \label{fig:resultofchannels}%
\end{figure}

According to Fig.~\ref{f:channel_3}, it can be clearly seen that there are only three features that are highly correlated to final classification of channel 03. By comparing (a) and (b) in Fig.~\ref{fig:resultofchannels} some of the features were selected for channel 03 that do not contribute to channel 05. This is why feature reduction must be performed before finalizing the feature vector which can be used to represent the raw sEMG signals. RMS, WA, MDF, SVD, MDF, RMS, RMS, ZCR, ZCR, WA are the finalized feature vector which includes two features from per channel which is ready for giving as an input to the classification model.

\subsection{Classification}

Since experiment is conducted for two activities, Non-linear support vector machine (SVM) one of SVMs~\cite{hsu2003practical} which is good for binary classification approaches was used to build final classification model. Behavior of non-linearity of extracted feature vector, radial basis function kernel (RBF kernel)~\cite{hsu2003practical} can be define as ${\displaystyle K(\mathbf {x} ,\mathbf {x'} )=\exp(-\gamma \|\mathbf {x} -\mathbf {x'} \|^{2})}$, where two input feature vectors denoted as x and x',  gamma can be represent as ${\displaystyle \textstyle \gamma ={\tfrac {1}{2\sigma ^{2}}}}$ where $\sigma$ is a free parameter, is used so that it will perform better in a high dimensional spaces. In other words, by using RBF kernels it will create more complex feature combinations by constructing hyperplanes among the classes in most satisfactory way.

The whole dataset is divided into two separate sets: training (8\%) and evaluation (2\%). Then to build a classification model five-fold cross-validation is employed on the training dataset. The final model was evaluated using evaluation dataset. Even though there were not much data to train classification model, training accuracy of sEMG classification is reported as $90\mypm3\%$ and evaluation accuracy is reported as $82\mypm4\%$. 

\section{Conclusions}\label{sec:Conclude}

MDTW approach based technique is proposed to automate the sEMG signals labelling process for given motions of interest in a near real-time manner. Label data validation is done by the part of the proposed algorithm itself by applying one to one DTW between separated out segments and corresponding motion of interest. Also, a number of repetitive actions for a particular action are known, this algorithm is able to extract most suitable segments for a given motion of interest with its distance vector. By changing allowed maximum distance result can be changed such that it won't detect any false positives.

Our future works will focus on implementing a deep neural network to detect repetitive patterns in a time series data without any prior-knowledge of motion of interests. 


 \bibliographystyle{ieeetr}
\bibliography{conference_041816}

\begin{thebibliography}{10}

\bibitem{frigo2009multichannel}
C.~Frigo and P.~Crenna, ``Multichannel semg in clinical gait analysis: a review
  and state-of-the-art,'' {\em Clinical Biomechanics}, vol.~24, no.~3,
  pp.~236--245, 2009.

\bibitem{solomonow1994surface}
M.~Solomonow, R.~Baratta, M.~Bernardi, B.~Zhou, Y.~Lu, M.~Zhu, and S.~Acierno,
  ``Surface and wire emg crosstalk in neighbouring muscles,'' {\em Journal of
  Electromyography and Kinesiology}, vol.~4, no.~3, pp.~131--142, 1994.

\bibitem{benatti2014analysis}
S.~Benatti, E.~Farella, E.~Gruppioni, and L.~Benini, ``Analysis of robust
  implementation of an emg pattern recognition based control.,'' in {\em
  BIOSIGNALS}, pp.~45--54, 2014.

\bibitem{bai2016shoulder}
D.~Bai, C.~Xia, J.~Yang, S.~Zhang, Y.~Jiang, and H.~Yokoi, ``Shoulder joint
  control method for smart prosthetic arm based on surface emg recognition,''
  in {\em Information and Automation (ICIA), 2016 IEEE International Conference
  on}, pp.~1267--1272, IEEE, 2016.

\bibitem{lynxmotional5d}
R.~inc, ``Lynxmotion - al5d,'' 2017.

\bibitem{de1997use}
C.~J. De~Luca, ``The use of surface electromyography in biomechanics,'' {\em
  Journal of applied biomechanics}, vol.~13, no.~2, pp.~135--163, 1997.

\bibitem{blana2016feasibility}
D.~Blana, T.~Kyriacou, J.~M. Lambrecht, and E.~K. Chadwick, ``Feasibility of
  using combined emg and kinematic signals for prosthesis control: A simulation
  study using a virtual reality environment,'' {\em Journal of Electromyography
  and Kinesiology}, vol.~29, pp.~21--27, 2016.

\bibitem{kinect}
Microsoft, ``Kinect setup xbox 360,'' 2017.

\bibitem{scherer2012kinect}
R.~Scherer, J.~Wagner, G.~Moitzi, and G.~M{\"u}ller-Putz, ``Kinect-based
  detection of self-paced hand movements: enhancing functional brain mapping
  paradigms,'' in {\em Engineering in Medicine and Biology Society (EMBC), 2012
  Annual International Conference of the IEEE}, pp.~4748--4751, IEEE, 2012.

\bibitem{wang2012human}
B.~Wang, C.~Yang, and Q.~Xie, ``Human-machine interfaces based on emg and
  kinect applied to teleoperation of a mobile humanoid robot,'' in {\em
  Intelligent Control and Automation (WCICA), 2012 10th World Congress on},
  pp.~3903--3908, IEEE, 2012.

\bibitem{frey:hal-01278245}
J.~Frey, ``{Comparison of a consumer grade EEG amplifier with medical grade
  equipment in BCI applications},'' in {\em {International BCI meeting}},
  (Asilomar, United States), May 2016.

\bibitem{vautard1992singular}
R.~Vautard, P.~Yiou, and M.~Ghil, ``Singular-spectrum analysis: A toolkit for
  short, noisy chaotic signals,'' {\em Physica D: Nonlinear Phenomena},
  vol.~58, no.~1, pp.~95--126, 1992.

\bibitem{golyandina2013singular}
N.~Golyandina and A.~Zhigljavsky, {\em Singular Spectrum Analysis for time
  series}.
\newblock Springer Science \& Business Media, 2013.

\bibitem{hassani2007singular}
H.~Hassani, ``Singular spectrum analysis: methodology and comparison,'' 2007.

\bibitem{spyrou2014singular}
L.~Spyrou, Y.~Blokland, J.~Farquhar, and J.~Bruhn, ``Singular spectrum analysis
  as a preprocessing filtering step for fnirs brain computer interfaces,'' in
  {\em Signal Processing Conference (EUSIPCO), 2014 Proceedings of the 22nd
  European}, pp.~46--50, IEEE, 2014.

\bibitem{berndt1994using}
D.~J. Berndt and J.~Clifford, ``Using dynamic time warping to find patterns in
  time series.,'' in {\em KDD workshop}, vol.~10, pp.~359--370, Seattle, WA,
  1994.

\bibitem{chowdhury2013surface}
R.~H. Chowdhury, M.~B. Reaz, M.~A. B.~M. Ali, A.~A. Bakar, K.~Chellappan, and
  T.~G. Chang, ``Surface electromyography signal processing and classification
  techniques,'' {\em Sensors}, vol.~13, no.~9, pp.~12431--12466, 2013.

\bibitem{daud2013features}
W.~M. B.~W. Daud, A.~B. Yahya, C.~S. Horng, M.~F. Sulaima, and R.~Sudirman,
  ``Features extraction of electromyography signals in time domain on biceps
  brachii muscle,'' {\em International Journal of Modeling and Optimization},
  vol.~3, no.~6, p.~515, 2013.

\bibitem{al2016distance}
H.~M. Al-Angari, G.~Kanitz, S.~Tarantino, and C.~Cipriani, ``Distance and
  mutual information methods for emg feature and channel subset selection for
  classification of hand movements,'' {\em Biomedical Signal Processing and
  Control}, vol.~27, pp.~24--31, 2016.

\bibitem{hermens1992median}
H.~Hermens, T.~Bruggen, C.~Baten, W.~Rutten, and H.~Boom, ``The median
  frequency of the surface emg power spectrum in relation to motor unit firing
  and action potential properties,'' {\em Journal of Electromyography and
  Kinesiology}, vol.~2, no.~1, pp.~15--25, 1992.

\bibitem{mahaphonchaikul2010emg}
K.~Mahaphonchaikul, D.~Sueaseenak, C.~Pintavirooj, M.~Sangworasil, and
  S.~Tungjitkusolmun, ``Emg signal feature extraction based on wavelet
  transform,'' in {\em Electrical Engineering/Electronics Computer
  Telecommunications and Information Technology (ECTI-CON), 2010 International
  Conference on}, pp.~327--331, IEEE, 2010.

\bibitem{vakkuri2004time}
A.~Vakkuri, A.~Yli-Hankala, P.~Talja, S.~Mustola, H.~Tolvanen-Laakso,
  T.~Sampson, and H.~Vierti{\"o}-Oja, ``Time-frequency balanced spectral
  entropy as a measure of anesthetic drug effect in central nervous system
  during sevoflurane, propofol, and thiopental anesthesia,'' {\em Acta
  Anaesthesiologica Scandinavica}, vol.~48, no.~2, pp.~145--153, 2004.

\bibitem{zhao2006emg}
J.~Zhao, Z.~Xie, L.~Jiang, H.~Cai, H.~Liu, and G.~Hirzinger, ``Emg control for
  a five-fingered underactuated prosthetic hand based on wavelet transform and
  sample entropy,'' in {\em Intelligent Robots and Systems, 2006 IEEE/RSJ
  International Conference on}, pp.~3215--3220, IEEE, 2006.

\bibitem{izenman2013linear}
A.~J. Izenman, ``Linear discriminant analysis,'' in {\em Modern multivariate
  statistical techniques}, pp.~237--280, Springer, 2013.

\bibitem{hsu2003practical}
C.-W. Hsu, C.-C. Chang, C.-J. Lin, {\em et~al.}, ``A practical guide to support
  vector classification,'' 2003.

\end{thebibliography}
%

\end{document}